\documentclass{article}
\usepackage{titling}

\pretitle{\begin{flushleft}\LARGE\bfseries}
\posttitle{\par\end{flushleft}}
\preauthor{\begin{flushleft}\large}
\postauthor{\par\end{flushleft}}
\predate{\begin{flushleft}}
\postdate{\par\end{flushleft}\vskip 0.5em}
\usepackage[svgnames]{xcolor}

\usepackage[customcolors,shade]{hf-tikz}
\usepackage{empheq,multicol,titlesec}
\usepackage[most]{tcolorbox}
\newtcbox{\mymath}[1][]{%
    nobeforeafter, math upper, tcbox raise base,
    enhanced, colframe=blue!30!black,
    colback=gray!30, boxrule=1pt,
    #1}
\usepackage[utf8]{inputenc} % allow utf-8 input
\usepackage[T1]{fontenc}    % use 8-bit T1 fonts
\usepackage{hyperref,amsmath,amsthm,amssymb,comment,tabularx}       % hyperlinks
\usepackage{url}            % simple URL typesetting
\usepackage{booktabs}       % professional-quality tables
\usepackage{amsfonts}       % blackboard math symbols
\usepackage{nicefrac}       % compact symbols for 1/2, etc.
\usepackage{microtype}      % microtypography
\usepackage[linesnumbered,ruled]{algorithm2e}
\usepackage{graphicx,caption,paralist,color,wrapfig}
\newtheorem{definition}{Definition}
\newtheorem{proposition}{Proposition}

\newcommand\numberthis{\addtocounter{equation}{1}\tag{\theequation}}
\newcommand{\argmin}{\operatornamewithlimits{argmin}}

\usepackage{multirow}
\usepackage{colortbl, array}
\usepackage{hhline}

\newcommand{\highest}[1]{\textcolor{Maroon}{\mathbf{#1}}}

\definecolor{rulecolor}{RGB}{0,71,171}
\definecolor{tableheadcolor}{RGB}{204,229,255}

\newcommand{\topline}{ %
        \arrayrulecolor{rulecolor}\specialrule{0.1em}{\abovetopsep}{0pt}%
        \arrayrulecolor{tableheadcolor}\specialrule{\belowrulesep}{0pt}{0pt}%
        \arrayrulecolor{rulecolor}}
\newcommand{\midtopline}{ %
        \arrayrulecolor{tableheadcolor}\specialrule{\aboverulesep}{0pt}{0pt}%
        \arrayrulecolor{rulecolor}\specialrule{\lightrulewidth}{0pt}{0pt}%
        \arrayrulecolor{white}\specialrule{\belowrulesep}{0pt}{0pt}%
        \arrayrulecolor{rulecolor}}
\newcommand{\bottomline}{ %
        \arrayrulecolor{white}\specialrule{\aboverulesep}{0pt}{0pt}%
        \arrayrulecolor{rulecolor} %
        \specialrule{\heavyrulewidth}{0pt}{\belowbottomsep}}%

\title{A Statistical Recurrent Model on the Manifold of Symmetric Positive Definite Matrices\thanks{This research was funded in part by the NSF grant IIS-1525431 and IIS-1724174
    to BCV and NSF CAREER award 1252725 and R01 EB022883 to VS. XZ and VS were also supported by UW CPCP (U54 AI117924).}}

\author{Rudrasis~Chakraborty$^\dag$ $\quad$ Chun-Hao~Yang$^{\dag\sharp}$ $\quad$ Xingjian~Zhen$^{\ddag\sharp}$ $\quad$ Monami~Banerjee$^\dag$ \quad Derek~Archer$^\dag$ $\quad$ David~Vaillancourt$^\dag$ $\quad$Vikas~Singh$^{\ddag}$ $\quad$ Baba~C.~Vemuri$^\dag$\\*[3pt] $^\dag$ University of Florida, Gainesville, USA \\ $^{\ddag}$ University of Wisconsin Madison, USA \\ {\tiny $^{\sharp}$ Equal contribution}}
\date{}

\begin{document}

\maketitle
\begin{abstract}
In a number of disciplines, the data (e.g., graphs, manifolds) to be
analyzed are non-Euclidean in nature.  Geometric deep learning
corresponds to techniques that generalize deep neural network models
to such non-Euclidean spaces. Several recent papers have shown how
convolutional neural networks (CNNs) can be extended to learn with
graph-based data.  In this work, we study the setting where the data
(or measurements) are ordered, longitudinal or temporal in nature and
live on a Riemannian manifold -- this setting is common in a variety
of problems in statistical machine learning, vision and medical
imaging. We show how recurrent statistical recurrent network models
can be defined in such spaces. We give an efficient algorithm and
conduct a rigorous analysis of its statistical properties. We perform
extensive numerical experiments demonstrating competitive performance
with state of the art methods but with significantly less number of
parameters. We also show applications to a statistical analysis task
in brain imaging, a regime where deep neural network models have only
been utilized in limited ways.
\end{abstract}

\section{Introduction}\label{intro}

%% ``Deep learning'' refers to learning complicated patterns present in the data in a hierarchical or multi-layer manner. While
In the last decade or so, deep neural network models have been
enormously successful in learning complicated patterns from data such
as images, videos and speech
\cite{lecun1998gradient,krizhevsky2012imagenet} -- this has led to a
number of technological breakthroughs as well as deployments in
turnkey applications. One of the popular neural network architectures
that has contributed to these advancements is convolutional neural
networks (CNNs). In the classical definition of convolution, one often
assumes that the data correspond to discrete measurements, acquired at
equally spaced intervals (i.e., Euclidean space), of a scalar (or
vector) valued function.
%on a discrete lattice grid 
Clearly, for images, the Euclidean lattice grid assumption makes sense
and the use of convolutional architectures is appropriate -- as
described in \cite{bronstein2017geometric}, a number of properties
such as stationarity, locality and compositionality follow.  While the
assumption that the underlying data satisfies the Euclidean structure
is explicit or implicit in an overwhelming majority of models,
%
%in which the underlying data has Euclidean structure,
recently there has been a growing interest in applying or extending
deep learning models for non-Euclidean data. This line of work is
called {\it Geometric deep learning} and typically deals with data
such as manifolds and graphs \cite{bronstein2017geometric}.  Existing
results describe strategies for leveraging the mathematical properties
of such geometric or structured data, specifically, lack of
\begin{inparaenum}[\bfseries (a)]
\item global linear structure, 
\item global coordinate system,
\item shift invariance/equivariance, 
\end{inparaenum}
by incorporating these ideas explicitly into deep networks used to
model them
\cite{chakraborty2018h,kondor2018generalization,cohen2018spherical,huang2016building,huang2017riemannian,cohen2016steerable}.

%% Such kind of applications can appear either as {\it functions on non-Euclidean space}
%%  or as {\it sample points on non-Euclidean space} \cite{}.
%As aptly stated in \cite{bronstein2017geometric}, {\it ``Geometric deep learning is an
%umbrella term for emerging techniques attempting to generalize (structured) deep neural
%models to non-Euclidean domains such as graphs and manifolds''}. The the main reasons to develop new
%machineries to apply deep learning on the non-Euclidean spaces are (i) lack of global linear structure
%(ii) lack of global coordinate system and (iii) lack of shift-invariance. 

Separate from the evolving body of work at the interface of
convolutional neural networks and structured data, there is a mature
literature in statistical machine learning
\cite{lebanon2015riemannian} and computer vision demonstrating how
exploiting the {\it structure} (or geometry) of the data can yield
advantages. Structured data abound in various data analysis tasks:
%%see hyunwoo dissertation, page 1. 
directional data in measurements from antennas
\cite{mammasis2010spherical}, time series data (curves) in finance
\cite{tsay2005analysis} and health sciences \cite{dominici2002use},
surface normal vectors on the unit sphere (in vision or graphics) \cite{straub2015dirichlet}, probability density functions
(in functional data analysis) \cite{srivastava2007riemannian},
covariance matrices (for use in conditional independences, image
textures) \cite{tuzel2006region}, rigid motions
(registration) \cite{park1995bezier}, shape representations (shape
space analysis) \cite{kendall1984shape}, tree-based data (parse trees
in natural language processing) \cite{quirk2005dependency}, subspaces
(videos, segmentation) \cite{xu2013gosus,elhamifar2009sparse},
low-rank matrices \cite{candes2009exact,vandereycken2013low}, and
kernel matrices \cite{scholkopf2002learning} are common examples.  In
neuroimaging, an image may have a structured measurement at each
voxel to describe water diffusion
\cite{basser1994mr,wang2005dti,Lenglet2006,Jian_NI07,aganj2009odf,cheng2012efficient}
or local structural change \cite{hua2008tensor,zacur2014multivariate,kim2017riemannian}.
And the study of the interface between geometry/structure and analysis
methods has given effective practical tools
%% --- for
%% instance, in medical imaging applications, analysis of diffusion
%% weighted Magnetic Resonance images where manifold-valued features such
%% as the diffusion tensors that are symmetric positive definite ($\textsf{SPD}$)
%% matrices that capture the diffusional behavior of water molecules at
%% each image voxel may be inferred from the raw diffusion MRI
%data. But,
because in order to define loss functions that make sense for the data
at hand, one needs to first define a metric which is intrinsic to the
structure of the data.
%% Naturally, the statistical tools such as
%% computing mean, variance, principal components, regression should be generalized to exploit on non-Euclidean data/ manifolds \cite{chakraborty2017statistics,salehian2013recursive,kim2014multivariate,chakraborty2016efficient}. 

%In contrast to
The foregoing discussion, for the most part, covers differential
geometry inspired algorithms for {\em non-sequential} (or
non-temporal) data.  The study of analogous schemes for temporal or
longitudinal data is less well-developed.
%exist for thelongitudinalregime where outside of[35, 17, 40, 27, 25], the literature remains sparse.
But analysis of dynamical scenes and stochastic processes is an
important area of machine learning and vision, and it is here that
some results have shown the benefits of explicitly using geometric
ideas. Some of the examples include the modeling of temporal evolution
of features in dynamic scenes in action recognition
\cite{afsari2012group,bissacco2001recognition,turaga2008statistical},
tractography \cite{cheng2015tractography,pujol2015dti} and so on.
There are also proposals describing modeling stochastic linear
dynamical system (LDS)
\cite{doretto2003dynamic,afsari2012group,bissacco2001recognition,turaga2008statistical}. In
\cite{afsari2012group,afsari2013alignment}, authors studied the
Riemannian geometry of LDS to define distances and first order
statistic.  Given that the marriage between deep learning and learning
on non-Euclidean domains is a fairly recent, the existing body of work
primarily deals with attempts to generalize the popular CNN
architectures.  Few if any results exist that study {\em recurrent} models
for non-Euclidean structured domains.

The broad success of Recurrent Neural Network (RNN) architectures
including Long short term memory (LSTM) \cite{hochreiter1997long} and
Gated recurrent unit (GRU) \cite{cho2014properties} in sequential
modeling like Natural Language Processing (NLP) has motivated a number
of attempts to apply such ideas to model stochastic processes or to
characterize dynamical scenes which can be viewed as a sequence of
images. Several works have proposed variants of RNN to model dynamical
scenes including \cite{srivastava2015unsupervised, donahue2015long,
  ng2015beyond, sharma2015action,yang2017tensor}. In the recent past,
developments have been made to reduce the number of parameters in RNN
and making RNN faster \cite{koutnik2014clockwork,yang2017tensor}. In
\cite{arjovsky2016unitary,henaff2016recurrent}, authors proposed an
efficient way to handle vanishing and exploding gradient problem of
RNN using unitary weight matrices. In \cite{jing2017gated}, authors
proposed a RNN model which combines the remembering ability of unitary
RNNs with the ability of gated RNNs to effectively forget redundant/
irrelevant information. Despite these results, we find that no
existing model describes a recurrent model for structured
(specifically, manifold-valued data). The {\bf main contribution} of
this paper is to describe a recurrent model (and accompanying theoretical
analysis) that will fall under the umbrella of ``geometric deep
learning'' --- it exploits the geometry of non-Euclidean data but is
specifically designed for temporal or ordered measurements.

\definecolor{violet}{HTML}{EE82EE}
\definecolor{lightblue}{HTML}{ADD8E6}
\definecolor{darkblue}{HTML}{55AAFF}
\definecolor{lightyellow}{HTML}{FFFFE0}
\definecolor{lightorange}{HTML}{FFAA7F}
\definecolor{darkorange}{HTML}{FF8C00}
\definecolor{lightgreen}{HTML}{90EE90}
\definecolor{darkgreen}{HTML}{538A53}

\section{Preliminaries: Key Ingredients from Riemannian geometry}
\label{prelim}
In this section, we will first give a brief overview of the Riemannian
geometry of $n \times n$ symmetric positive definite matrices
(henceforth will be denoted by $\textsf{SPD}(n)$). Note that our
development is not limited to $\textsf{SPD}(n)$, but choosing a
specific manifold simplifies the presentation and the notations
significantly. Then, we will present key ingredients needed for our
proposed recurrent model.

\vspace{0.05cm}
\textbf{Differential Geometry of $\textsf{SPD}(n)$:} Let $\textsf{SPD}(n)$
be the set of $n\times n$ symmetric positive definite matrices. The
group of $n\times n$ full rank matrices, denoted by $\textsf{GL}(n)$
and called the general linear group, acts on $\textsf{SPD}(n)$ via the group
action, $g. A := gAg^T$, where $g \in \textsf{GL}(n)$ and $A \in \textsf{SPD}(n)$. One can define a $\textsf{GL}(n)$ invariant intrinsic metric,
$d_{\textsf{GL}}$ on $\textsf{SPD}(n)$ as follows \cite{helgason1962differential} $d_{\textsf{GL}}(A,B) =
\sqrt{\textsf{trace}(\textsf{Log}(A^{-1}B)^2)}$. Here, $\textsf{Log}$ is the matrix logarithm. This metric is
intrinsic but requires a spectral decomposition for computation, a
computationally intensive task for large matrices. In
\cite{cherian2011efficient}, the Jensen-Bregman LogDet (JBLD) divergence was
introduced on $\textsf{SPD}(n)$. As the name suggests, this is not a
metric but as proved in \cite{sra2011positive}, the square root of
JBLD is however a metric (called the Stein metric), which is
defined as $d(A,B) = \sqrt{\log \text{det} (\frac{A+B}{2})
- \frac{1}{2} \log \text{det}(AB)}$.

Here, we used the notation $d$ without any subscript to denote the
Stein metric. It is easy to see that the Stein metric is
computationally much more efficient than the
$\textsf{GL}(n)$-invariant natural metric on $\textsf{SPD}(n)$ as no
eigen decomposition is required. This will be very useful for training
our recurrent model. In the rest of the paper, we will assume the
metric on $\textsf{SPD}(n)$ to be the Stein metric. Now, we will
describe a few operations on $\textsf{SPD}(n)$ which are needed to
define the recurrent model.

\vspace{0.05cm}
\textbf{``Translation'' operation on $\textsf{SPD}(n)$:} Let $I$ be the 
set of all isometries on $\textsf{SPD}(n)$, i.e., given $g \in I$,
$d(g.A, g.B) = d(A, B)$, for all $A, B \in \textsf{SPD}(n)$, where $.$
is the group action as defined earlier. It is clear that $I$ forms a
group (henceforth, will be denoted by $G$) and for a given $g \in G$
and $A \in \textsf{SPD}(n)$, $g.A \mapsto B$, for some
$B \in \textsf{SPD}(n)$ is a group action. One can easily see that,
endowed with the Stein metric, $G = \textsf{GL}(n)$. In this work, we will choose a subgroup of $\textsf{GL}(n)$, i.e., $\textsf{O}(n)$ as our choice of $G$, where,  $\textsf{O}(n)$ is the set of $n \times n$
orthogonal matrices and $g.A := gAg^T$.
As the $\textsf{O}(n)$ group operation preserves the distance, we call
this group operation ``translation'', analogous to the case of
Euclidean space and is denoted by $\textsf{T}_A(g) := gAg^T$.

\vspace{0.05cm}
\textbf{Parametrization of $\textsf{SPD}(n)$:} Let 
$A \in \textsf{SPD}(n)$. We will obtain the Cholesky factorization of
$A = LL^T$, where $L$ is an invertible lower traingular matrix. This
gives a unique parametrization of $\textsf{SPD}(n)$. Let the
parametrization be $A = \textsf{Chol}((l_1, l_2, \cdots l_n,\cdots,
l_{n(n+1)/2})^t)$. With a slight abuse of notation, we will use
$\textsf{Chol}$ to denote both decomposition and construction based on
the type of the domain of the function, i.e., $\textsf{Chol}(A) := L$
and $\textsf{Chol}(L) := LL^T = A$. Note that here $l_1, l_2, \cdots,
l_n$ are diagonal entries of $L$ and are positive and
$l_{n+1}, \cdots, l_{n(n+1)/2}$ can be any real numbers.

\vspace{0.05cm}
{\bf Parametrization of $\textsf{O}(n)$}: $\textsf{O}(n)$ is a Lie
group \cite{hall2015lie} of $n \times n$ orthogonal matrices (of
dimension $n(n-1)/2$) with the corresponding Lie algebra,
$\mathfrak{O}(n)$, and consists of the set of $n \times n$
skew-symmetric matrices. The Lie algebra is a vector space, so we will
use the corresponding element from the Lie algebra to parametrize a
point on $\textsf{O}(n)$. Let $g \in \textsf{O}(n)$, we will use the
matrix logarithm of $\mathfrak{g} = \textsf{log}(g)$ to get the
parametrization as the skew-symmetric matrix. So, $g
= \textsf{exp}((\mathfrak{g}_1, \mathfrak{g}_2, \cdots, \mathfrak{g}_{n(n-1)/2})^t)$. $\textsf{exp}$
is the matrix exponential operator.

\vspace{0.05cm}
{\bf Weighted Fr\'{e}chet mean (wFM) of matrices on
$\textsf{SPD}(n)$:} Given
$\left\{X_i\right\}_{i=1}^N \subset \textsf{SPD}(n)$, and
$\left\{w_i\right\}_{i=1}^N$ with $w_i \geq 0$, for all $i$ and
$\sum_i w_i = 1$, the weighted Fr\'{e}chet mean
(wFM) \cite{frechet1948elements} is:
\setlength{\abovedisplayskip}{1pt}
\setlength{\belowdisplayskip}{1pt}
\begin{align}
\label{theory:eq1}
M^* = \argmin_{M} \sum_{i=1}^N w_i d^2\left(X_i, M\right)
\end{align}

The existence and uniqueness of the Fr\'{e}chet mean (FM) is discussed
in detail in \cite{afsari2011riemannian}. For the rest of the paper,
we will assume that the samples lie within a geodesic ball of
appropriate radius so that FM exists and is unique. We will use
$\textsf{FM}(\left\{X_i\right\}, \left\{w_i\right\})$ to denote the
wFM of $\left\{X_i\right\}$ with weights $\left\{w_i\right\}$.
With the above tools in hand, now we are ready to formulate the
Statistical Recurrent Neural Network on $\textsf{SPD}(n)$, dubbed as
SPD-SRU.

\section{Statistical Recurrent Network Models in the 
space of $\textsf{SPD}(n)$ matrices}
\label{theory}

%We will present a recurrent model on $\textsf{SPD}(n)$.
The main motivation for our work comes from the statistical recurrent
unit (SRU) model on Euclidean spaces in \cite{oliva2017statistical}.
To setup our formulation, we will briefly review the SRU formulation
followed by details of our recurrent model for manifold valued
measurements.

\paragraph{What is the Statistical Recurrent Unit (SRU)?}
 
The authors in \cite{oliva2017statistical} propose an interesting
model for sequential (or temporal) data based on an un-gated recurrent
unit (called Statistical Recurrent Unit (SRU)). The model maintains
the sequential dependency in the input samples through a simple
summary statistic --- the so-called exponential moving average.  Even
though the proposal is based on an un-gated architecture, the
development and experiments show that the results from SRU are
competitive with more complex alternatives like LSTM and GRU. One
reason put forth in that work is that using appropriately designed
summary statistics, one can essentially emulate complicated gated
units and still capture long terms relations (or memory) in sequences.
This property is particularly attractive when we study recurrent
models for more complicated measurements such as manifolds.  Recall
that the key challenge in extending statistical machine learning
models to manifolds involves re-deriving many of the classical
(Euclidean) arithmetic and geometric operations while respecting the
geometry of the manifold of interest. The simplicity of un-gated units
provides an excellent starting point.
%without compromising performance motivates us to generalize a recurrent model on Riemannian manifolds based on summary statistics.
Below, we describe the key update equations that define the SRU.

%% \begin{empheq}[box=\mymath]{equation*}
%%     c_i = \langle\psi|\phi\rangle
%% \end{empheq}

\begin{wrapfigure}{r}{0.47\textwidth}
\centering
\includegraphics[scale=0.40]{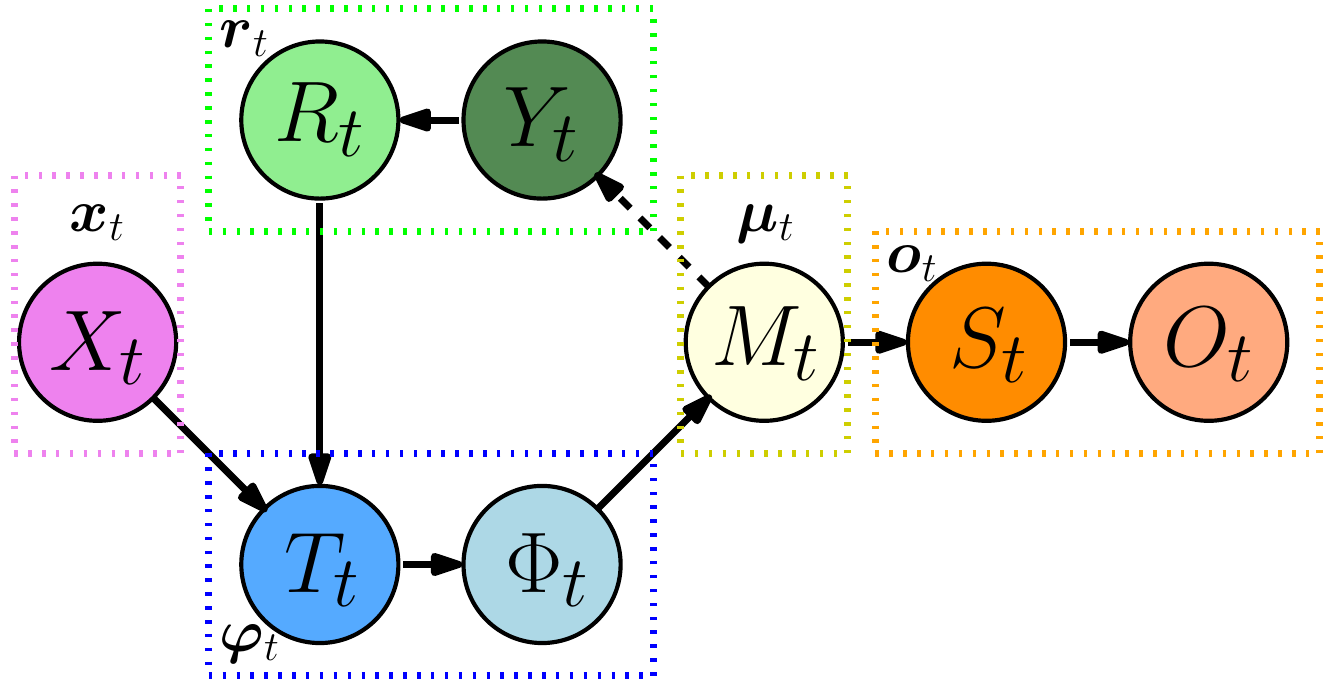}
\caption{\footnotesize Sketch of an SPD-SRU and SRU layer (dashed line represnets dependence on the previous time point).}
\label{theory:fig1}
\end{wrapfigure}

Let $\mathbf{x}_1, \mathbf{x}_2, \cdots \mathbf{x}_T$ be an input
sequence on $\mathbf{R}^n$, presented to the model. As in most
recurrent models, the training process in SRU proceeds by updating the
weights of the model.  Let the weight matrix be denoted by $W$ (the
node is indexed by the superscript).  The update rules for SRU are as
follows:
\begin{minipage}{0.47\textwidth}
\begin{small}
\begin{align}
\hfsetfillcolor{white}
\hfsetbordercolor{green}
\tikzmarkin{a}(4.4,-0.25)(-0.2,0.4)
&\mathbf{r}_t = \text{ReLU}\left(W^{(r)}\boldsymbol{\mu}_{t-1}+b^{(r)}\right) \label{theory:eq21-s}\\
\hfsetfillcolor{white}
\hfsetbordercolor{blue}
\tikzmarkin{a}(5.6,-0.25)(-0.2,0.4)
&\boldsymbol{\varphi}_t = \text{ReLU}\left(W^{(\phi)}\mathbf{r}_t+W^{(x)}\mathbf{x}_t+b^{(\phi)}\right) \label{theory:eq22-s}
\end{align} 
\end{small}
\end{minipage}
\begin{minipage}{0.52\textwidth}
\begin{small}
\begin{align}
\hfsetfillcolor{white}
\hfsetbordercolor{yellow}
\tikzmarkin{a}(5.4,-0.22)(-0.2,0.4)
\forall \alpha \in J \ , \quad \boldsymbol{\mu}_{t}^{(\alpha)} = \alpha\boldsymbol{\mu}_{t-1}^{(\alpha)} + (1-\alpha)\boldsymbol{\varphi}_t \label{theory:eq23-s}&\\
\hfsetfillcolor{white}
\hfsetbordercolor{orange}
\tikzmarkin{a}(4.05,-0.27)(-0.2,0.33)
\mathbf{o}_t = \text{ReLU}\left(W^{(o)}\boldsymbol{\mu}_t+b^{(o)}\right) \label{theory:eq24-s}&
\end{align}
\end{small}
\end{minipage}
where $J$ is the set of different scales. The SRU formulation is
analogous to mean map embedding (MME) but applied to non
i.i.d. samples.  Since the average of a set of i.i.d. samples will
essentially marginalize over time, simple averaging will lose the
temporal/sequential information.  On the other hand, the SRU computes
a moving average over time which captures the average of the data seen
so far, i.e., when computing $\boldsymbol{\mu}$ from
$\boldsymbol{\varphi}$ (as shown in Fig. \ref{theory:fig1}). This is
very similar to taking the average of stochastic processes and looking
at the ``average process''.  Further, by looking at averages over
different scales, essentially, we can uncover statistics computed over
different time scales.  This is because $\boldsymbol{\mu}$ is not only
a function of $\boldsymbol{\phi}$ but also a function of
$\left\{\mathbf{x}_i\right\}_{i=1}^{t-1}$ via $\mathbf{r}_t$.  This
dependence on the past ``tokens'' in the sequence is shown in
Fig. \ref{theory:fig1} by a ``dashed'' line.  With this description,
we can easily list the key operational components in the update rules
in \eqref{theory:eq21-s}-\eqref{theory:eq24-s} and then evaluate if
such components can be generalized to serve as the
%then present a generalization of each of the ingredients, which will be the
building blocks of our proposed model.

\vspace{0.1cm}
{\bf Which low-level operations are needed?} We can verify that
the key ingredients to define the model in SRU are 
\begin{inparaenum}[\bfseries (i)] \item weighted sum; \item
    addition of bias; \item moving average and \item
    non-linearity.  
\end{inparaenum} 
In principle, if we can generalize each of these operations to the
$\textsf{SPD}(n)$ manifold, it will provide us the basic components to
define the model.  Observe that items (i) and (iii) are essentially a
{\bf weighted sum} if we impose a convexity constraint on the weights.
Then, the weighted sum for the Euclidean setting can be generalized
using wFM as defined in Section \ref{prelim} (denoted by
$\textsf{FM}$).
    
\setlength{\intextsep}{0pt}%
\setlength{\columnsep}{4pt}%
    
If we can do so, it will also provide a way to compute moving averages
on $\textsf{SPD}(n)$.  Now, the second operation we can identify above
is the {\bf translation} on Euclidean spaces. This can be achieved by
the ``translation'' operation on $\textsf{SPD}(n)$ as defined in
Section \ref{prelim} (denoted by $\textsf{T}$).  Finally, in order to
generalize ReLU on $\textsf{SPD}(n)$, we will use the standard ReLU on
the parameter space (this will be the local chart of
$\textsf{SPD}(n)$) and then map it back on to the manifold.  This
means that we have generalized each of the key components. With this
in hand, we are ready to present our proposed recurrent model on
$\textsf{SPD}(n)$. We first formally describe our SPD-SRU layer and
then contrast with the SRU layer, to help see the main differences.

\paragraph{Basic components of the {SPD-SRU model}.}
Let, $X_1, X_2, \cdots X_T$ be an input temporal or ordered sequence
of points on $\textsf{SPD}(n)$. The update rules for a layer of
SPD-SRU is as
follows: 
\begin{small} 
\begin{align} \hfsetfillcolor{darkgreen} \hfsetbordercolor{darkgreen!20!white} \tikzmarkin{a}(4.4,-0.25)(-0.2,0.4)
Y_t
= \textsf{FM}\left(\left\{M_{t-1}^{(\alpha)}\right\}, \left\{w^{(y,\alpha)}\right\}\right) \
, \quad& \hfsetfillcolor{lightgreen} \hfsetbordercolor{lightgreen!20!white} \tikzmarkin{a}(2.4,-0.25)(-0.2,0.4)
R_t = \textsf{T}\left(Y_t,
g^{(r)}\right) \label{theory:eq21}\\ \hfsetfillcolor{darkblue} \hfsetbordercolor{darkblue!20!white} \tikzmarkin{a}(3.65,-0.25)(-0.2,0.4)
T_t = \textsf{FM}\left(\left\{R_t, X_t\right\}, w^{(t)}\right) \
, \quad& \hfsetfillcolor{lightblue} \hfsetbordercolor{lightblue!20!white} \tikzmarkin{a}(2.4,-0.25)(-0.2,0.4) \Phi_t
= \textsf{T}\left(T_t,
g^{(p)}\right) \label{theory:eq22}\\ \hfsetfillcolor{lightyellow} \hfsetbordercolor{lightyellow!20!white} \tikzmarkin{a}(1.0,-0.25)(-0.2,0.4) \forall \alpha \in
J \ , \quad
& \hfsetfillcolor{lightyellow} \hfsetbordercolor{lightyellow!20!white} \tikzmarkin{a}(4.3,-0.25)(-0.2,0.4)
M_t^{(\alpha)}
= \textsf{FM}\left(\left\{M_{t-1}^{(\alpha)}, \Phi_t\right\}, \alpha\right) \label{theory:eq23}\\ \hfsetfillcolor{darkorange} \hfsetbordercolor{darkorange!20!white} \tikzmarkin{a}(4.4,-0.25)(-0.2,0.4)
S_t
= \textsf{FM}\left(\left\{M_{t}^{(\alpha)}\right\}, \left\{w^{(s,\alpha)}\right\}\right) \
, \quad& \hfsetfillcolor{lightorange} \hfsetbordercolor{lightorange!20!white} \tikzmarkin{a}(6.0,-0.25)(-0.2,0.4)
O_t
= \textsf{Chol}\left(\text{ReLU}\left(\textsf{Chol}\left(\textsf{T}\left(S_t,
g^{(y)}\right)\right)\right)\right) \label{theory:eq24} \end{align} \end{small}
where, $t \in \left\{1, \cdots, T\right\}$ and $M_0^{(\alpha)}$ is
initialized to be a diagonal $n\times n$ matrix with small positive
values.  Similar to before, the set $J$ consists of positive real
numbers from the unit interval. Now, computing the FM at the different
elements of $J$ will give a wFM at different ``scales'', exactly as
desired.  Analogous to the SRU, here $M_t^{(\alpha)}$s are computed by
averaging $\Phi_t$ at different scales as shown in
Fig. \ref{theory:fig1}.  This model leverages the context based on
previous data by asking the moving averages, $M_t^{(\alpha)}$ to
depend on past
data, \begin{tiny}$\left\{X_i\right\}_{i=1}^{t-1}$\end{tiny} through
$R_t$ (as shown in Fig. \ref{theory:fig1}).

\vspace{0.1cm}
{\bf Comparison between the SPD-SRU and the SRU layer:} In the SPD-SRU
unit above, each update identity is a generalization of an update
equation of SRU.  In \eqref{theory:eq21}, we compute the weighted
combination of the previous FMs (computed using different ``scales'')
with a ``translation'', i.e., the input
is \begin{tiny}$\left\{M_{t-1}^{(\alpha)}\right\}$\end{tiny} and the
output is $R_t$.  This update equation is analogous to the weighted
combination of the past means with bias as given
in \eqref{theory:eq21-s}) where the input
is \begin{tiny}$\left\{\boldsymbol{\mu}_{t-1}^{(\alpha)}\right\}$\end{tiny}
and the output is $\mathbf{r}_t$. This update rule calculates a
weighted combination of the past information.  In \eqref{theory:eq22},
we compute a weighted combination of the previous information, $R_t$
and the current point or token, $X_t$ with a ``translation''.  The
input of this equation is $R_t$ and $X_t$ and the output is
$\Phi_t$. This is analogous to \eqref{theory:eq22-s}, where the input
is $\mathbf{r}_t$ and $\mathbf{x}_t$ and the output is
$\boldsymbol{\varphi}_t$. This update rule combines old and new
information.  Now, we will update the new information based on the
combined information at the current time step, i.e., $\Phi_t$.  This
is accomplished in \eqref{theory:eq23}. Here, we are computing an FM
(average) at different ``scales''.  Computing averages at different
``scales'' essentially allows including information from previous data
points which have been seen at various time scales.  This step is a
generalization of \eqref{theory:eq23-s}. In this step, the input is
$\left\{M_{t}^{(\alpha)}\right\}$ and $\Phi_t$
(with \begin{tiny}$\left\{\boldsymbol{\mu}_{t-1}^{(\alpha)}\right\}$\end{tiny}
and $\boldsymbol{\varphi}_t$ respectively) and the output
is \begin{tiny}$\left\{M_{t}^{(\alpha)}\right\}$\end{tiny}
(with \begin{tiny}$\left\{\boldsymbol{\mu}_{t}^{(\alpha)}\right\}$\end{tiny}). This
step is the combined information gathered at the current time step.
Finally, in \eqref{theory:eq24}, we used a weighted combination of the
current FMs (averages) and outputs $O_t$.  This is the last update
rule in SRU, i.e., \eqref{theory:eq24-s}. Observe that we did {\em
not} use the ReLU operation in each update rule of SPD-SRU, in
contrast to SRU. This is because, these update rules are highly
nonlinear unlike in the SRU, hence, a ReLU unit at the final output of
the layer is sufficient.  Also, notice that $O_t \in \textsf{SPD}(n)$,
hence, we can cascade multiple SPD-SRU layers, in other words in the
next layer, the input sequence will be $O_1, O_2 \cdots O_T$.  The
update equations track the ``averages'' (FM) at varying scales. This
is the reason we call our framework statistical recurrent network.  We
will shortly see that our framework can utilize parameters more
efficiently and requires very few parameters because of the ability to
use the covariance structure.

\vspace{0.1cm}
{\bf Important properties of SPD-SRU model:} The ``translation''
operator $\mathsf{T}$ is analogous to ``adding'' a bias term in a
standard neural network. One reason we call it ``translation'' is
because the action of $\textsf{O}(n)$, preserves the metric.  Notice
that although in this description, we track the FMs at different
scales, one may easily use other statistics, e.g., Fr\'{e}chet median
and mode, etc.  The key bottleneck is to efficiently compute the
moving statistic (whatever it may be), which will be discussed
shortly.  Note that the SPD-SRU formulation can be generalized to
other manifolds.  In fact, it can be easily generalized to Riemannian
homogeneous spaces \cite{helgason1962differential} because of two
reasons \begin{inparaenum}[\bfseries (a)] \item closed form
expressions for Riemannian exponential and inverse exponential maps
exist and \item a group $G$ acts transitively on these spaces, hence
we can generalize the definition of ``translation''.  \end{inparaenum}
Other manifolds are also possible but the technical details will be
different.  Now, we will comment on learning the parameters of our
proposed model.

\vspace{0.1cm}
{\bf Learning the parameters:} Notice that using the parametrization
of $\textsf{O}(n)$, we will learn the ``bias'' term on the parametric
space, which is a vector space. The weights in the wFM must satisfy
the non-negativity constraint. In order to ensure that this property
is satisfied, we will learn the square root of the weights which is
unconstrained, i.e., the entire real line.  We will impose the affine
constraint explicitly by normalizing the weights. Hence, all the
trainable parameters lie in the Euclidean space and the optimization
of these parameters is unconstrained, hence standard techniques are
sufficient.

{\bf Remarks.} It is interesting to observe that the update equations
in \eqref{theory:eq21}-\eqref{theory:eq24} involve group operations
and wFM computation.  But as evident from the \eqref{theory:eq1}, the
wFM computation requires numerical optimization, which is
computationally {\em not} efficient.  This is a bottleneck. For
example, for our proposed model, on a batch size of $20$ with
$15\times 15$ matrices with $T=50$, we need to compute FM $3000$
times, even for just $10$ epochs.  Next, we will develop formulations
to make this wFM computation faster since it is called hundreds of
times in a typical training procedure.
       %In the next section, we will focus our attention to develop a novel update rule for the weighted FM computation. 

\section{An efficient way to compute the wFM on $\textsf{SPD}(n)$}
\label{efficientfm}
%And as evident from \eqref{theory:eq1}, computation of
The foregoing discussion describes how the computation of wFM needs an
optimization on the $\textsf{SPD}$ manifold.  If this sub-module is
slow, the demands of the overall runtime will rule out practical
adoption.  In contrast, if this sub-module is fast but numerically or
statistically unstable, the errors will propagate in unpredictable
ways, and can adversely affect the parameter estimation.  Thus, we
need a scheme that balances performance and efficiency.

Estimation of the FM from samples is a well researched topic.  For
instance, the authors
in \cite{moakher2006symmetric,pennec2006riemannian} used Riemannian
gradient descent to compute the FM. But the algorithm
has a runtime complexity of $\mathcal{O}(iN)$, where $N$ is the number
of samples and $i$ is the number of iterations for convergence. This
procedure comes with provable consistency guarantees -- thus,
while it will serve our goals in theory, we find that the runtime for
each run makes training incredibly slow.
%Because of the large time complexity of the gradient descent based method, though this method gives an FM estimator which is consistent, it is not appropriate for our RNN model.
%We came across the paper by Salehian et al. in \cite{salehian2013recursive},
On the other hand, the $\mathcal{O}(N)$ recursive FM estimator using
the Stein metric presented in \cite{salehian2013recursive} is fast and
apt for this task if no additional assumptions are made.
%where the authors proposed a recursive formulation to compute FM using Stein metric. This formulation takes $\omega(N)$ time, and as this is the minimum time complexity to
%compute FM, it fits perfectly in our formulation. But unfortunately, this formulation though it's very fast, does not have any
However, it comes with no theoretical guarantees of consistency.
%In other words, the proposed FM estimator in \cite{salehian2013recursive} may diverge.
%As we need nested FM computation a lot of times in our RNN formulation, without the proof of consistency, this FM estimator is not useful as the result may diverge.
%As a result, extensive iterative usage of such a module may
%require elaborate wrapper mechanisms to ensure correct performance.
%So, we have two options either to use a consistent estimator which is time consuming and make it fast or to use a fast algorithm which is not consistent.
\paragraph{Key Observation.} 
%While our initial attempts focused on 
%making the schemes in \cite{moakher2006symmetric,pennec2006riemannian}
%more efficient, 
We found that with a few important changes to the idea
described in \cite{salehian2013recursive}, one can derive an FM
estimator that retains the attractive efficiency behavior and
is provably consistent. The key ingredient here involves using
%We went ahead with the second option and came up with
a novel isometric mapping from the $\textsf{SPD}$ manifold to the unit
Hilbert sphere. 
%By slightly adjusting the calculations, the new scheme
%turns out to have a number of interesting technical properties. 
Next, we present the main idea followed by the analysis.

\paragraph{Proposed Idea.}
Let $\left\{X_i\right\}_{i=1}^N \subset \textsf{SPD}(n)$ for which we
want to compute the FM which will be used
in \eqref{theory:eq21}--\eqref{theory:eq24}.  Authors
in \cite{salehian2013recursive} presented a recursive Stein mean estimator
given below:
\setlength{\abovedisplayskip}{1pt}
\setlength{\belowdisplayskip}{1pt}
\begin{align*}
  \label{theory:eq3.5}
M_1 = X_1 \, \quad M_k = M_{k-1}\left[\sqrt{T_k + \frac{(2w_k-1)^2}{4}\left(I-T_k\right)^2}-\frac{2w_k-1}{2}\left(I - T_k\right)\right], \numberthis
\end{align*}
where $T_k = M_{k-1}^{-1}X_k$ and $\left\{w_i\right\}$ is the set of
weights. 
%the consistency we are seeking for. Below, we first present the method proposed in \cite{salehian2013recursive} and then develop
Instead, briefly, our strategy is 
\begin{inparaenum}[\bfseries (i)]
\item use an isometric mapping from $\textsf{SPD}(n)$ to the unit
Hilbert sphere; 
\item make use of an efficient way to compute the FM on the unit
Hilbert sphere; 
\end{inparaenum}
%Let $\left\{X_i\right\}_{i=1}^N$ be the set of samples on $\textsf{SPD}(n)$. In \cite{salehian2013recursive},
%the authors  derived
This isometric mapping to the Hilbert sphere then transfers the
problem of proving consistency of the estimator from $\textsf{SPD}(n)$
to that on the Hilbert sphere, which is easier to prove as shown
below.  This then leads to consistency of FM estimator on
$\textsf{SPD}(n)$.

We define the isometric mapping from $\textsf{SPD}(n)$ with a Stein
metric to $\mathbf{S}^{\infty}$, i.e., the infinite dimensional unit
hypersphere. In order to define it, notice that we need to define a
metric, $d_S$ on $\mathbf{S}^{\infty}$ such that, $(\textsf{SPD}(n),
d)$ and $(\mathbf{S}^{\infty}, d_S)$ are {\em isometric}. This
procedure and the associated consistency analysis is described below
(all proofs are in the supplement).

\begin{definition}
\label{theory:def1}
Let $A \in \textsf{SPD}(n)$. Let $f:=\mathcal{G}(A)$ be the Gaussian
density with $\mathbf{0}$ mean and covariance matrix $A$. Now, we
normalize the density $f$ by $f \mapsto f/\|f\|$ to map it onto
$\mathbf{S}^{\infty}$. Let, $\Phi: \textsf{SPD}(n) \rightarrow
\mathbf{S}^{\infty}$ be that mapping. We define the metric on
$\mathbf{S}^{\infty}$ as \begin{small}$d_S(\widetilde{f}, \widetilde{g})
= \sqrt{-\log \langle \widetilde{f}, \widetilde{g} \rangle^2}$\end{small}.
\end{definition}

Here, $\langle,\rangle$ is the $L^2$ inner product.The following
proposition proves the isometry between $\textsf{SPD}(n)$ with the
Stein metric and the hypersphere with the new metric. Let, $A,
B \in \textsf{SPD}(n)$. Then,
\begin{proposition}
\label{theory:prop2}
Let $\widetilde{f} = \Phi(A)$ and
$\widetilde{g} = \Phi(B)$. Then, $d(2A, 2B) =
d_S(\widetilde{f}, \widetilde{g})$.
\end{proposition}
Note that, $\Phi$ maps a point on $\textsf{SPD}(n)$ to the positive
orthant of $\mathbf{S}^{\infty}$, denoted by $\mathcal{H}$ since the
components of any probability vector are non-negative.
We should point out that in this metric space, there are no geodesics
since it is {\em not} a length space. As a result, we cannot simply
use the consistency proof of the stochastic gradient descent based FM
estimator presented in \cite{bonnabel2013stochastic} for any
Riemannian manifold and apply it here.  Hence, the recursive FM
presented next for the identity in \eqref{theory:eq3.5} with the
mapping described above will need a separate consistency analysis.
%algorithm but also it's consistency proof.
%Because of the isometry property this will in turn prove the consistency of the estimator in \eqref{theory:eq3.5}.

%\vspace{0.2cm}
\paragraph{Recursive Fr\'{e}chet mean algorithm on $(\mathcal{H}, d_S)$.} 
Let $\left\{\mathbf{x}_i\right\}_{i=1}^N$ be the samples on
$(\mathcal{H}, d_S)$ where $\mathcal{H}$ gives the positive orthant of
$\mathbf{S}^{\infty}$.  Then, the FM of the given samples, denoted by
$\mathbf{m}^*$, is defined as $\mathbf{m}^*
= \arg\min_{\mathbf{m}} \sum_{i=1}^N d_S^2
(\mathbf{x}_i, \mathbf{m})$. Our recursive algorithm to compute the wFM of
$\left\{\mathbf{x}_i\right\}_{i=1}^N$ is:

\setlength{\abovedisplayskip}{1pt}
\setlength{\belowdisplayskip}{1pt}
\begin{align*}
\label{theory:eq4}
\mathbf{m}_1 = \mathbf{x}_1 \, \quad \mathbf{m}_k = \arg \min_{\mathbf{x}} \left(w_k\; d^2(\mathbf{x}_{k}, \mathbf{x}) + (1-w_k)\; d^2(\mathbf{m}_{k-1}, \mathbf{x})\right) \numberthis 
\end{align*}
where, $\mathbf{m}_k$ is the $k^{th}$ estimate of the FM. At each step
of our algorithm, we simply calculate a wFM of two points and we chose
the weights to be the Euclidean weights. So, in order to construct a
recursive algorithm, we need to have a closed form expression of the
wFM, as stated next.

\begin{proposition}
\label{theory:prop4}
The minimizer of \eqref{theory:eq4} is given by \begin{small}$\mathbf{m}_k =
\frac{\sin(\theta-\alpha)}{\sin(\theta)} \mathbf{m}_{k-1} +
\frac{\sin(\alpha)}{\sin(\theta)} \mathbf{x}_k$\end{small}, where $\theta =
\arccos(\langle \mathbf{m}_{k-1}, \mathbf{x}_k\rangle )$ and\begin{small} $\alpha =
\arctan\left(\frac{-1+\sqrt{4c^2(1-w_k)-4c^2(1-w_k)^2+1}}{2c(1-w_k)}\right)$\end{small} and $c = \tan(\theta)$.
\end{proposition}

\paragraph{Consistency and Convergence analysis of the estimator.}
The following proposition (see supplement for proof) gives us the weak
consistency of this estimator and also the convergence rate.

\begin{proposition}
\label{theory:prop5}
\begin{inparaenum}[\bfseries (a)]
\item $\text{Var}\left(\mathbf{m}_k\right) \rightarrow 0$ as $k \rightarrow \infty$.
\item The rate of convergence of the proposed recursive FM estimator 
is super linear.
\end{inparaenum}
\end{proposition}
Due to proposition \ref{theory:prop2}, we obtain a consistency result
for \eqref{theory:eq3.5} with our mapping.  These results suggest that
we now have a suitable FM estimator which is {\bf consistent and
efficient} -- this can be used as a black-box module in our RNN
formulation in \eqref{theory:eq21}-\eqref{theory:eq24}.
%Furthermore, we also have a {\it novel mapping from $\textsf{SPD}$ to $\mathbf{S}^{\infty}$} and {\it a novel recursive fast consistent FM estimator on $\mathbf{S}^{\infty}$}.

%Furthermore, we also have a {\it novel mapping from $\textsf{SPD}$ to $\mathbf{S}^{\infty}$} and {\it a novel recursive fast consistent FM estimator on $\mathbf{S}^{\infty}$}.

\section{Experiments}\label{results}
In this section, we demonstrate the application of SPD-SRU to answer
three important questions \textbf{(1)} Using the manifold constraint,
what are we saving in terms of number of parameters/ time and is the
performance competitive? \textbf{(2)} When data is manifold valued,
can we still use our framework with the geometry constraint?  \textbf{(3)}
In a real application, how much improvements can we get over the
baseline?

We perform three sets of experiments to answer these questions namely:
(1) classification of moving patterns on Moving MNIST data, (2)
classification of actions on UCF11 data and (3) permutation testing to
detect group differences between patients with and without Parkinson's
disease. In the following subsections, we discuss about each of these
dataset in more detail and present the performance of our SPD-SRU. Our code is available at \url{https://github.com/zhenxingjian/SPD-SRU/tree/master}.
%% Following paragraph is duplicated at the beginning of the next
%% subsection. I am therefore taking it out. 

%%We compare with LSTM
%% \cite{hochreiter1997long}, SRU \cite{oliva2017statistical}, TT-GRU
%% and TT-LSTM \cite{yang2017tensor}. In the first two classification
%% applications, we use a convolution block (with $c$ channels) before
%% SPD-SRU for all the competitive methods except TT-RNN. Before the
%% SPD-SRU layer, we include a covariance block analogous to
%% \cite{yu2017second} after one convolution layer (see
%% \cite{yu2017second} for details of the covariance block
%% construction). So, the input of SPD-SRU is a sequence of matrices
%% in $\textsf{SPD}(c+1)$.
\vspace{0.2cm}
%on a unit sphere shown in Fig. \ref{exp:fig3}.
\subsection{Savings in terms of number of parameters/ time 
and experiments on vision datasets.} In this section, we perform two sets of
experiments namely (1) classification of moving patterns on Moving
MNIST data, (2) classification of actions on UCF11 data to show the
improvement of our proposed framework over the state-of-the-art
methods in terms of number of parameters/ time.  We compared with LSTM
\cite{hochreiter1997long}, SRU \cite{oliva2017statistical}, TT-GRU and
TT-LSTM \cite{yang2017tensor}. In the first two classification
applications, we use a convolution block before the recurrent unit for
all the competitive methods except for TT-GRU and TT-LSTM.  In our
SPD-SRU model, before the recurrent layer, we included a covariance
block analogous to \cite{yu2017second} after one convolution layer
(\cite{yu2017second} includes details of the construction for the
covariance block). So, the input of our SPD-SRU layer is a sequence of
matrices in $\textsf{SPD}(c+1)$, where $c$ is the number of channels
from the convolution layer.
\vspace{0.1cm}
\paragraph{Classification of moving patterns on Moving MNIST data}

We used the Moving MNIST data as generated in
\cite{srivastava2015unsupervised}. For this experiment we did $2$ and
$3$ classes classification experiment. In each class, we generated
$1000$ sequences each of length $20$ showing $2$ digits moving in a
$64\times 64$ frame. Though within a class, the digits are random, we
fixed the moving pattern by fixing the speed and direction of the
movement. In this experiment, we kept the speed to be same for all the
sequences, but two sequences from two different classes can differ in
orientation by at least $5^\circ$ and by atmost $30^\circ$. We
experimentally see that, SPD-SRU can achieve very good $10$-fold
testing accuracy even when the orientation difference of two classes
is $5^\circ$. In fact SPD-SRU takes the smallest number of parameters
among all methods tested and still offers the best average testing
accuracy.

\setlength{\intextsep}{0pt}%
\setlength{\columnsep}{5pt}%
\begin{wraptable}{r}{0.8\textwidth}
   \centering
   \scalebox{0.75}{
\begin{tabular}{cccccc} 
\topline\rowcolor{tableheadcolor}
 & & {\bf time (s)} & \multicolumn{3}{c}{{\bf orientation ($^\circ$)}} \\
\arrayrulecolor{tableheadcolor}\hhline{---~~~}\arrayrulecolor{rulecolor}\hhline{~~~---}\rowcolor{tableheadcolor}
\multirow{-2}{*}{\bf Mode} & \multirow{-2}{*}{\bf \# params.} & {\bf / epoch} & $30$-$60$ & $10$-$15$ & $10$-$15$-$20$ \\
\midtopline
SPD-SRU & $\highest{1559}$ & $\sim 6.2$ & $\highest{1.00\pm 0.00}$ & $\highest{0.96 \pm 0.02}$ & $\highest{0.94 \pm 0.02}$ \\
TT-GRU & $2240$ & $\sim\highest{2.0}$ & $\highest{1.00 \pm 0.00}$ & $0.52 \pm 0.04$ & $0.47 \pm 0.03$ \\
TT-LSTM & $2304$ & $\sim\highest{2.0}$ & $\highest{1.00 \pm 0.00}$ & $0.51 \pm 0.04$ & $0.37 \pm 0.02$ \\
SRU & $159862$ & $\sim 3.5$ & $\highest{1.00 \pm 000}$ & $0.75 \pm 0.19$ & $0.73 \pm 0.14$ \\
LSTM & $252342$ & $\sim 4.5$ & $0.97 \pm 0.01$ & $0.71 \pm 0.07$ & $0.57 \pm 0.13$ \\
\bottomline
\end{tabular}
}
\caption{\footnotesize Comparative results on Moving MNIST}
\label{results:tab1}
\end{wraptable}

In Table \ref{results:tab1}, we report the mean and standard deviation
of the $10$-fold testing accuracy. We like to point out that the
training accuracy for all the competitive methods is $>95\%$ for all
cases. For TT-RNN, we reshaped the input to be $4\times 8 \times 8
\times 16$ and kept the output shape and rank to be $4\times 4 \times
4 \times 4$ and $1 \times 4 \times 4 \times 4 \times 1$. The number of
output units for LSTM is kept as $10$ and the number of statistics for
SRU is kept as $80$. Note that, we chose different parameters for SRU
and LSTM and TT-RNN and the one we reported here are the one for which
the number of parameters are smallest for the reported testing
accuracy. For the convolution layer, we chose the kernel size to be
$5\times 5$ and the input and output channels to be $5$ and $10$
respectively, i.e., the dimension of the $\textsf{SPD}$ matrix is $11$
for this experiment. As before, the parameters are chosen so the
number of parameters are smallest to get the reported testing
accuracy.

\begin{wrapfigure}{r}{0.4\textwidth}
\centering
\includegraphics[scale=0.30]{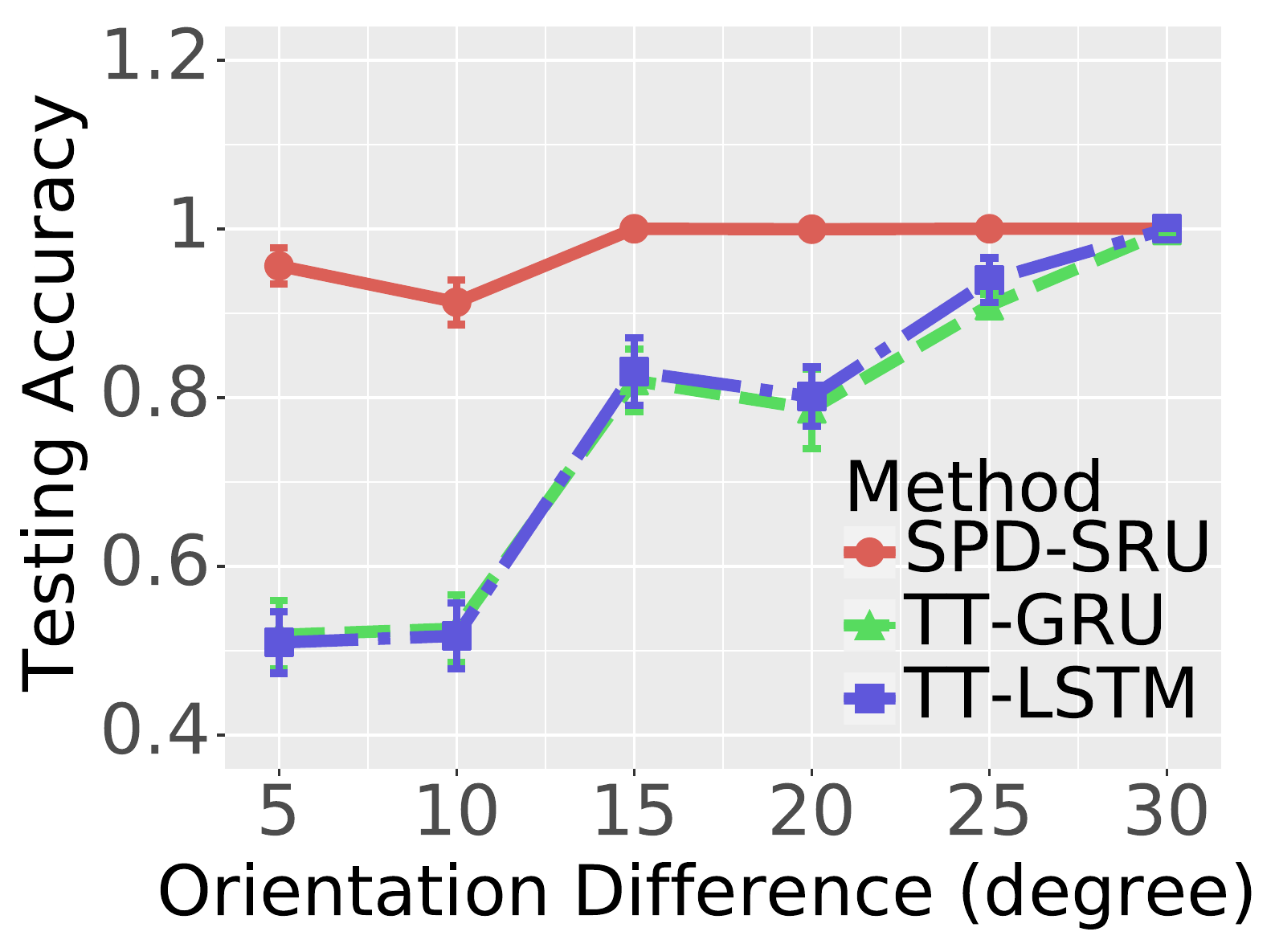}
\caption{\footnotesize Comparison of testing accuracies with varying orientations}
\label{results:fig1}
\end{wrapfigure}

One can see from the table that, SPD-SRU takes the least number of
parameters and can achieve very good classification accuracy even for
$5^{\circ}$ orientation difference and for three classes.  Note that
TT-RNN is the closest to SPD-SRU in terms of parameters.
%and surprisingly the test accuracy of that methods
%suffers for $5^{\circ}$ orientation difference,
For comparisons, we do another experiment where we vary the difference
of orientation from $30^{\circ}$ to $5^{\circ}$.  The testing
accuracies are shown in Fig. \ref{results:fig1}. We can see that only
SPD-SRU maintains good $10$-fold testing accuracy for all orientation
differences while the performance of TT-RNN (both variants)
deteriorates as we decrease the difference between orientations of the
two classes (the effect size).  

\setlength{\intextsep}{0pt}%
\setlength{\columnsep}{5pt}%

In terms of training time, SPD-SRU
takes around $6$ seconds per epoch while the fastest method is TT-RNN
which takes around $2$ seconds. But, in this experiment, SPD-SRU takes
$75$ epochs to converge to the reported results while TT-RNN takes
around $400$ epochs. So, although TT-RNN is faster per epoch, the
total training time for TT-RNN and SPD-SRU are almost the same.  We
also like to point out that though the number of trainable parameters
are fewer for SPD-SRU than TT-RNN, the time difference is due to
constructing the covariance in each epoch which can be optimized via
faster implementations.
\vspace{0.1cm}
\paragraph{Classification of moving patterns on UCF-11 data}

We performed an action classification experiment on UCF11 dataset
\cite{liu2009recognizing}. It contains in total 1600 video clips
belonging to 11 classes that summarize the human action visible in
each video clip such as basketball shooting, diving and others.  We
followed the same processing step as done in
\cite{yang2017tensor}. Each frame has resolution $320 \times 240$. We
generate a sequence of RGB frames of size $160 \times 120$ from each
clip at $24$ fps. The lengths of frame sequences from each video
therefore are in the range of $204$-$1492$ with an average of
$483.7$. For SPD-SRU, we chose two convolution layers with kernel size
$7 \times 7$ and number of output channels to be $5$ and $7$
respectively and then $5$ PSRN layers. Hence, the dimension of the
covariance matrices are $8\times 8$ for this experiment. For TT-GRU
and TT-LSTM, we used the same configurations of input and output
factorization as given in \cite{yang2017tensor}. For SRU and LSTM we
used the number of statistics and number of output units to be
$750$. For both SRU and LSTM we used $3$ convolution layers with
kernel size $7 \times 7$ and output channels to be $10$, $15$ and $25$
respectively to get the reported testing accuracies. All the models
achieve $>90\%$ training accuracy. We report the testing accuracy with
the number of parameters and time per epoch in Table
\ref{results:tab2}. From this experiment, we can see that the number
of parameters for SPD-SRU is significantly smaller than the other
models without sacrificing the testing accuracy. In terms of training
time, SPD-SRU takes approximately $3$ times more time than TT-RNN but
SPD-SRU (TT-RNN) converges in $50$ ($100$) epochs. Furthermore, we
like to point out that after $400$ epochs, SPD-SRU gives $79.90\%$
testing accuracy. Hence, analogous to the previous experiment, we can
conclude that SPD-SRU maintains very good classification accuracy
while keeping the number of trainable parameters very
small. Furthermore, this experiment indicates that SPD-SRU can achieve
competitive performance on real data with small number of training
parameters in comparable time.

\vspace{0.2cm}
\subsection{Application on manifold  valued data}
From the previous two experiments, we can conclude that SPD-SRU
requires a smaller number of parameters. In this subsection, we focus
our attention to a neuroimaging application where data is manifold
valued. Because the number of parameters are small, we can do
statistical testing on brain connectivity at the fiber bundle
level. We seek to find group differences between subjects with and
without Parkinson's disease (denoted by `PD' and `CON') based on the
M1 fiber tracts on both hemispheres of the brain.

\setlength{\intextsep}{0pt}%
\setlength{\columnsep}{5pt}%
\begin{wraptable}{r}{0.58\textwidth}
   \centering
   \scalebox{0.75}{
\begin{tabular}{cccc} 
\topline\rowcolor{tableheadcolor}
{\bf Model} & {\bf \# params.} & {\bf time/ epoch} & {\bf Test acc.}  \\
\midtopline
SPD-SRU & $\highest{3337}$ & $\sim 76$ & $\highest{0.78}$ \\
TT-GRU & $6048$ & $\sim 42$ & $\highest{0.78}$ \\
TT-LSTM & $6176$ & $\sim\highest{33}$ & $\highest{0.78}$ \\
SRU & $2535630$ & $\sim 50$ & $0.75$ \\
LSTM & $14626425$ & $\sim 57$ & $0.70$ \\
\bottomline
  \end{tabular}
}
\caption{\footnotesize Comparative results on UCF11 data}
\label{results:tab2}
\end{wraptable}

\begin{wrapfigure}{r}{0.30\textwidth}
\centering
\includegraphics[scale=0.20]{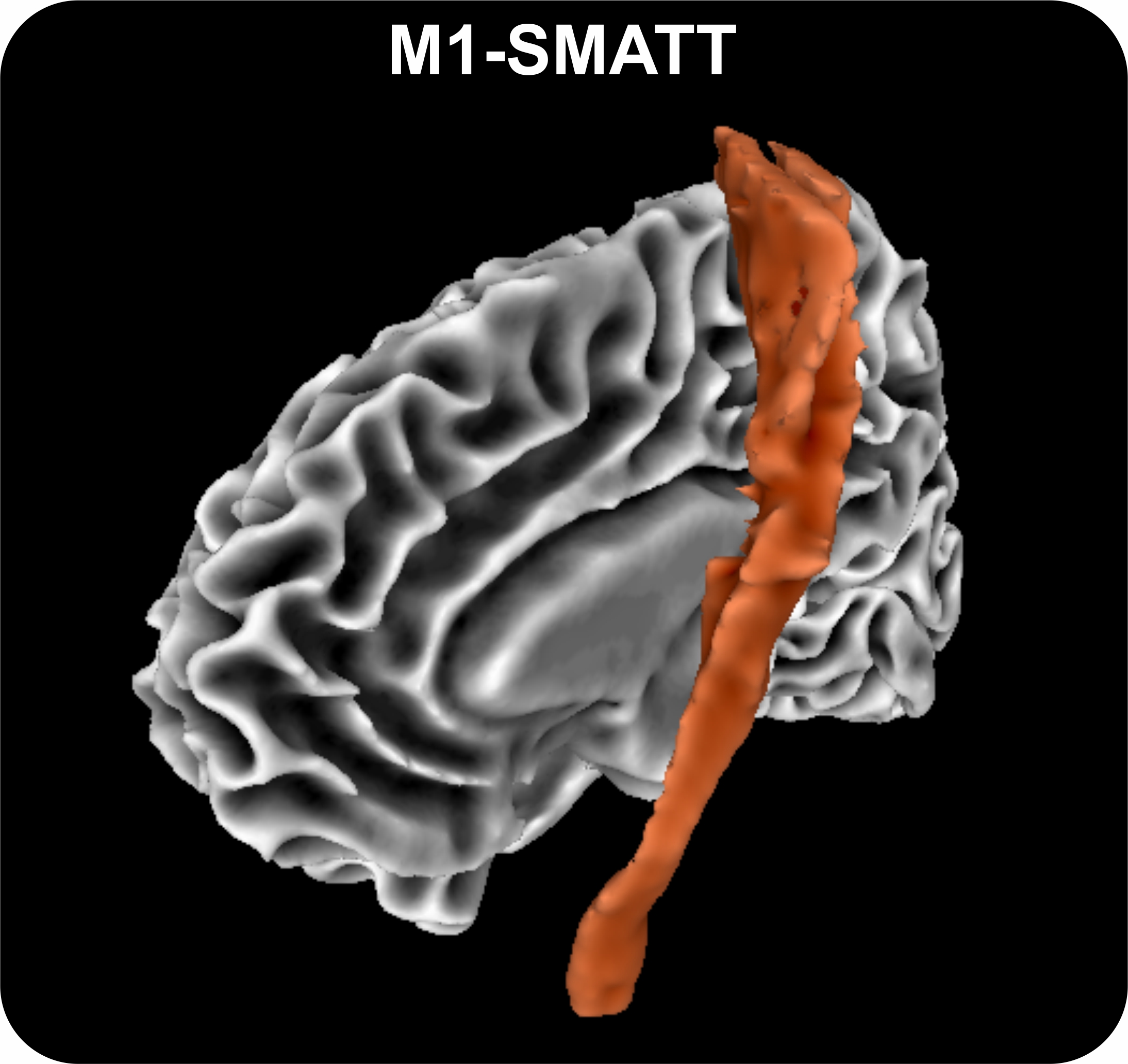}
\caption{\footnotesize M1-SMATT template}
\label{results:fig2}
\end{wrapfigure}

\vspace{0.1cm}
\paragraph{Permutation testing to detect group differences}
The data pool consists of dMRI (human) brain scans acquired from $50$
`PD' patients and $44$ `CON'. All images were collected using a 3.0 T
MR scanner (Philips Achieva) and 32-channel quadrature volume head
coil.  The parameters of the diffusion imaging acquisition sequence
were as follows: gradient directions = 64, b-values = 0/1000 s/mm2,
repetition time =7748 ms, echo time = 86 ms, flip angle =
$90^{\circ}$, field of view = $224 \times 224$ mm, matrix size = $112
\times 112$, number of contiguous axial slices = 60 and SENSE factor P
= 2. We used FSL \cite{behrens2007probabilistic} software to extract
M1 fiber tracts (denoted by `LM1' and `RM1')
\cite{archer2017template}, which consists of $33$ and $34$ points
respectively (please see Fig. \ref{results:fig2} for M1-SMATT fiber
tract template). We fit a diffusion tensor and extract $3\times 3$
$\textsf{SPD}$ matrices. Now, for each of these two classes, we use
$3$ layers of SPD-SRU to learn the tracts pattern to get two models
for `PD' and `CON' (denoted by `mPD' and `mCON').

\setlength{\intextsep}{0pt}%
\setlength{\columnsep}{5pt}%

Now, we use a permutation testing based on a ``distance'' between
`mPD' and `mCON'. We will define the distance between two network
models as proposed in \cite{triacca2016measuring} (let it be denoted
by $d_{\text{mod}}$). Here, we assume each subject is independent
hence use of permutation testing is sensible. Then we perform
permutation testing for each tract as follows (i) randomly permute the
class labels of the subjects and learn `mPD' and `mCON' models for
each of the new group. (ii) compute $d^j_{\text{mod}}$ (iii) repeat
step (ii) 10,000 times and report the $p$-value as the fraction of
times $d^j_{\text{mod}} > d_{\text{mod}}$. So, we ask if we can
reject the null hypothesis that {\it there is no significant
difference between the tracts models learned from the two different
classes}.

As a baseline, we use the following scheme: (i) for each tract of each
subject, compute the FM of the matrices on the tract. (ii) use
Cramer's test based on this Stein distance. (iii) do the permutation
testing based on the Cramer's test.
 
We found that using our SPD-SRU model with $3$ layers, the $p$-value
for `LM1' and `RM1' are $0.01$ and $0.032$ respectively, while the
baseline method gives a p-value of $0.17$ and $0.34$
respectively. Hence, we conclude that, unlike the baseline method,
using SPD-SRU we can reject the null hypothesis with $95\%$
confidence.  To the best of our knowledge, this is the first result
that demonstrates a RNN based statistical significance test applied on
tract based group testing in neuroimaging.

\section{Conclusions}
Non-Euclidean or manifold valued data are ubiquitous in science and
engineering.  In this work, we study the setting where the data (or
measurements) are ordered, longitudinal or temporal in nature and live
on a Riemannian manifold. This setting is common in a variety of
problems in statistical machine learning, vision and medical
imaging. We presented a generalization of the RNN to such
non-Euclidean spaces and analyze its theoretical properties.  Our
proposed framework is fast and needs far fewer parameters than the
state-of-the-art.  Extensive experiments show competitive performance
on benchmark computer vision data in comparable time.  We also apply
our framework to perform statistical analysis in brain connectivity
and demonstrate the applicability to manifold valued data.

{\small
\bibliographystyle{plain}
\bibliography{references}
}

\end{document}